%% file: main.tex
\documentclass[letterpaper, 10 pt, conference]{ieeeconf}  
\IEEEoverridecommandlockouts                       

\usepackage{cite}
\usepackage{algorithmic}
\usepackage{graphicx}
\usepackage{mathrsfs}
\usepackage{textcomp}
\usepackage{xcolor}
\usepackage{comment}
\usepackage{amssymb}

\usepackage{amsmath, amsthm}

\usepackage{siunitx}
\usepackage{soul}
\usepackage{multirow}
\usepackage{subcaption}
\usepackage{siunitx}
\usepackage[font=footnotesize]{caption}
\usepackage[colorlinks=true,linkcolor=black,anchorcolor=black,citecolor=black,filecolor=black,menucolor=black,runcolor=black,urlcolor=black]{hyperref}
\usepackage{mathtools}
\usepackage[ruled,vlined,linesnumbered]{algorithm2e}
\usepackage{bm}
\usepackage[normalem]{ulem}
\usepackage{nomencl}
\usepackage{etoolbox}

\usepackage{amsfonts}

\def\lnext{\bigcirc}
\def\eventually{\Diamond}
\def\until{\mathcal{U}}
\def\true{\top}

\input{macros}

\title{\LARGE \bf
Terrain-Aware Model Predictive Control of Heterogeneous Bipedal and Aerial Robot Coordination for Search and Rescue Tasks}

\author{Abdulaziz Shamsah$^{1,2}$, Jesse Jiang$^{3}$, Ziwon Yoon$^{1}$, Samuel Coogan$^{3}$, and Ye Zhao$^{1}$
\thanks{$^{1}$Woodruff School of Mechanical Engineering, Georgia Institute of Technology, Atlanta, GA 30313, USA. (e-mail: ashamsah3@gatech.edu, zyoon6@gatech.edu, ye.zhao@me.gatech.edu)}
\thanks{$^{2}$Mechanical Engineering Department, College of Engineering and Petroleum, Kuwait University, PO Box 5969, Safat, 13060, Kuwait}
\thanks{$^3$School of Electrical and Computer Engineering, Georgia Institute
of Technology, Atlanta, GA 30332 USA (e-mail: jjiang@gatech.edu, sam.coogan@gatech.edu).}
}

\begin{document}

\maketitle
\thispagestyle{empty}
\pagestyle{empty}


\input{sections/00_abstract}
\input{sections/01_intro}

\input{sections/02_problem_formulation}

\input{sections/03_terrainGP}
\input{sections/04_taskplanner}
\input{sections/05_MPC}

\input{sections/07_results}
\input{sections/08_conclusion}



\bibliographystyle{IEEEtran}
\bibliography{main.bib}

\end{document}

%% file: macros.tex
\newtheorem{thm}{Theorem}[section]

\newtheorem{prob}[thm]{Problem}

\newtheorem{defn}{Definition}[section]

\newcommand{\set}[1]{\mathcal{#1}}

\newcommand{\state}{\boldsymbol{x}}

\newcommand{\ctrl}{\boldsymbol{u}}

\newcommand{\position}{\boldsymbol{p}}

\newcommand{\nextq}{_{q+1}}
\newcommand{\currq}{_{q}}

\newcommand{\local}{^{\rm loc}}

\newcommand{\lip}{^{\rm LIP}}
\newcommand{\lipone}{^{\rm LIP1}}
\newcommand{\liptwo}{^{\rm LIP2}}
\newcommand{\quadrotor}{^{\rm quad}}
\newcommand{\subject}{^{\rm s}}

%% file: sections/00_abstract.tex
\begin{abstract}
Humanoid robots offer significant advantages for search and rescue tasks, thanks to their capability to traverse rough terrains and perform transportation tasks. In this study, we present a task and motion planning framework for search and rescue operations using a heterogeneous robot team composed of humanoids and aerial robots. We propose a terrain-aware Model Predictive Controller (MPC) that incorporates terrain elevation gradients learned using Gaussian processes (GP). This terrain-aware MPC generates safe navigation paths for the bipedal robots to traverse rough terrain while minimizing terrain slopes, and it directs the quadrotors to perform aerial search and mapping tasks. The rescue subjects' locations are estimated by a target belief GP, which is updated online during the map exploration. A high-level planner for task allocation is designed by encoding the navigation tasks using syntactically cosafe Linear Temporal Logic (scLTL), and a consensus-based algorithm is designed for task assignment of individual robots. We evaluate the efficacy of our planning framework in simulation in an uncertain environment with various terrains and random rescue subject placements.

\end{abstract}

%% file: sections/01_intro.tex
\section{Introduction}

\begin{figure}[t]
\centerline{\includegraphics[width=.9\columnwidth]{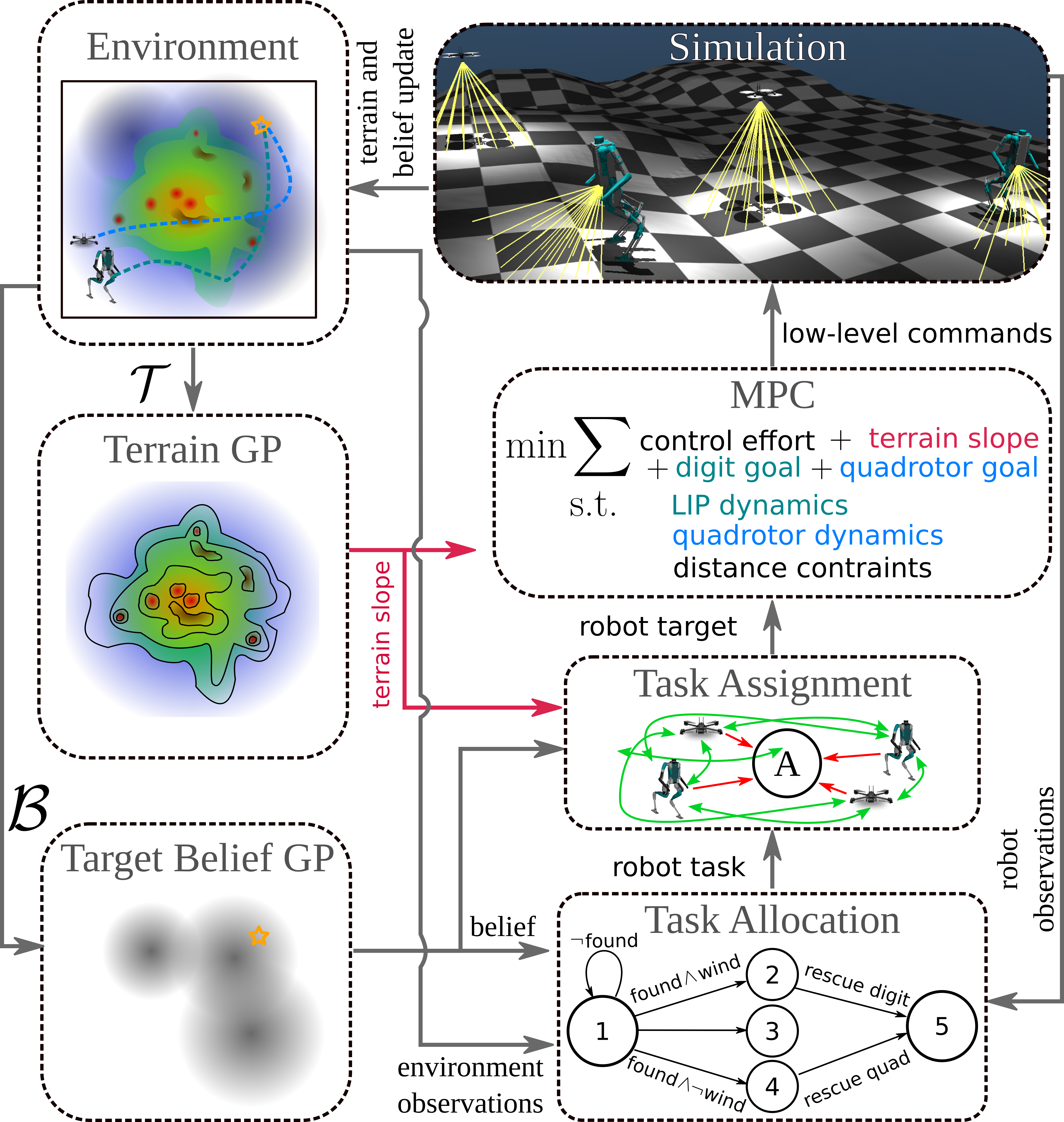}}
\caption{Overall block diagram of the proposed heterogeneous bipedal and aerial robot planner for search and rescue in terrain-uncertain environments.}
\label{fig:block_framework}
\vspace{-0.2in}
\end{figure}

Humanoid robots demonstrate unique capabilities to traverse over rough and sloped terrains~\cite{gibson2022terrain,mccrory2023bipedal,huang2023efficient,gu2024robust, shamsah2023integrated} while performing manipulation and payload transportation tasks~\cite{murooka2021humanoid,radosavovic2024real}. Such capabilities are essential for search and rescue missions. However, humanoids suffer from locomotion instability in extreme and uncertain environments, limited field of view for searching, and inferior agility compared to quadrotors~\cite{kaufmann2023champion,papachristos2017uncertainty}. To this end, we propose a multi-robot task and motion planning framework (see Fig.~\ref{fig:block_framework}) that leverages the capabilities of teamed-up bipedal and aerial robots for search and rescue missions. To our knowledge, this work presents the first framework that incorporates humanoids as part of a heterogeneous robot team in search and rescue tasks.

For humanoid locomotion, numerous studies have investigated navigation on sloped terrains~\cite{mccrory2023bipedal,huang2023efficient,gibson2022terrain, zhao2022reactive}, often focusing on the avoidance of sloped terrains or finding planar footholds. The work in~\cite{gibson2022terrain}, closely related to the work we propose here, uses a piecewise linear terrain approximation for computing foot placements for a linear inverted pendulum model~\cite{kajita20013d}. In~\cite{gibson2022terrain}, terrain information is assumed to be known or approximated by the operator, whereas in our work we employ a Gaussian process (GP)~\cite{Williams96gaussianprocesses, jiang2022safe} map to learn terrain elevation. Given this, our framework enables the Model Predictive Controller (MPC) to compute optimal paths that allow the humanoid to traverse sloped terrain while minimizing terrain slopes in a cost function. Such capability of traversing sloped terrain generalizes our previous works on integrating terrain GPs with locomotion, which have focused on avoiding uncertain or elevated terrains~\cite{jiang2023abstraction,muenprasitivej2024bipedal}.

Multi-robot task allocation (MRTA) has been widely explored in the community using centralized~\cite{quinton2023market} or decentralized~\cite{chakraa2023optimization, choi2009consensus} communication. Such an MRTA problem is commonly solved based on three main methods: market-based, behavior-based, and optimization-based approaches~\cite{chakraa2023optimization,khamis2015multi,quinton2023market}. To the best of our knowledge, there have been very few works 
leveraging the heterogeneous robot capabilities of humanoids and drones~\cite{parker2016exploiting,fu2022robust, silano2021power}. 
In the proposed heterogeneous robot teaming problem, task allocation requires matching of robot capabilities with assigned tasks, which makes market-based approaches ideal~\cite{badreldin2013comparative}. Therefore, we use consensus-based methods~\cite{choi2009consensus,chen2019distributed,ye2021decentralized}, as they strike a balance between optimality and efficiency. As opposed to a discretized task space, where conflict may arise from assigning the same task to multiple robots~\cite{choi2009consensus}, our task requires coordinated exploration of a continuous map.
To this end, we propose a new conflict resolution scheme tailored towards removing redundant map exploration. 

Formal control methods such as Linear Temporal Logic (LTL) and its variants have been commonly used to formally specify multi-robot tasks \cite{menghi2018multi, schillinger2017multi, kloetzer2016multi}. A major advantage of LTL-based planning approaches is that they give formal guarantees on task satisfaction. However, most LTL planning works consider scenarios in which the targeted locations are static and known \textit{a priori}. For scenarios in which target locations must be dynamically updated at runtime, such as what we consider in this work, the literature \cite{grover2021semantic} uses an exploration policy in concert with the task planner. 

In this study, we present a hierarchical approach to bipedal and aerial multi-robot motion planning for search and rescue (see Fig.~\ref{fig:block_framework}), integrating a task allocation planner at the high level (Sec~\ref{subsec:task_allocation}), consensus-based task assignment module at the middle level (Sec~\ref{sec:task_assignment}), and a multi-robot terrain-aware MPC at the low level (Sec.~\ref{sec:mpc}).
The core contributions are as follows: (1) Integration of a terrain GP into an MPC that solves the optimal paths for the multi-robot team while maximizing the traversability for the bipedal robots, (2) Consensus-based task assignment module that assigns targets to the heterogeneous multi-robot team to match the capabilities of each robot, and (3) an scLTL task allocation module that dynamically allocates high-level tasks to the robot team based on current low-level robot observations of the environment and the progress of task completion.

%% file: sections/02_problem_formulation.tex
\section{Problem formulation}\label{sec:prob_formulation}

\subsection{Task and Environment Setup}
We task a team of bipedal robots and quadrotors with a search and rescue mission in an environment with partially observable and uneven terrain. The task is to search for subjects and deliver them to a safe location known \textit{a priori}.

We assume that an initial set of terrain elevation data points $\set{T} = \{ (\position_1, e_1), (\position_2, e_2), \ldots, (\position_n, e_n) \}$ is known \textit{a priori}, where $\position=(x,y)$ is the 2-D position of the data point and $e$ is the corresponding terrain elevation. 
We have initial belief points $\set{B} = \{ (\position\subject_1, c_1\subject), (\position\subject_2, c_2\subject), \ldots, (\position\subject_m, c_m\subject) \}$ corresponding to the 2-D Euclidean positions of the subject location $\position\subject = (x\subject, y\subject)$ and the corresponding belief confidence level $c\subject$ that the subject is located at this point.

\subsection{Bipedal Robot Walking Model}
\label{subsec:ROM}
\begin{figure}[t]
\centerline{\includegraphics[width=.99\columnwidth]{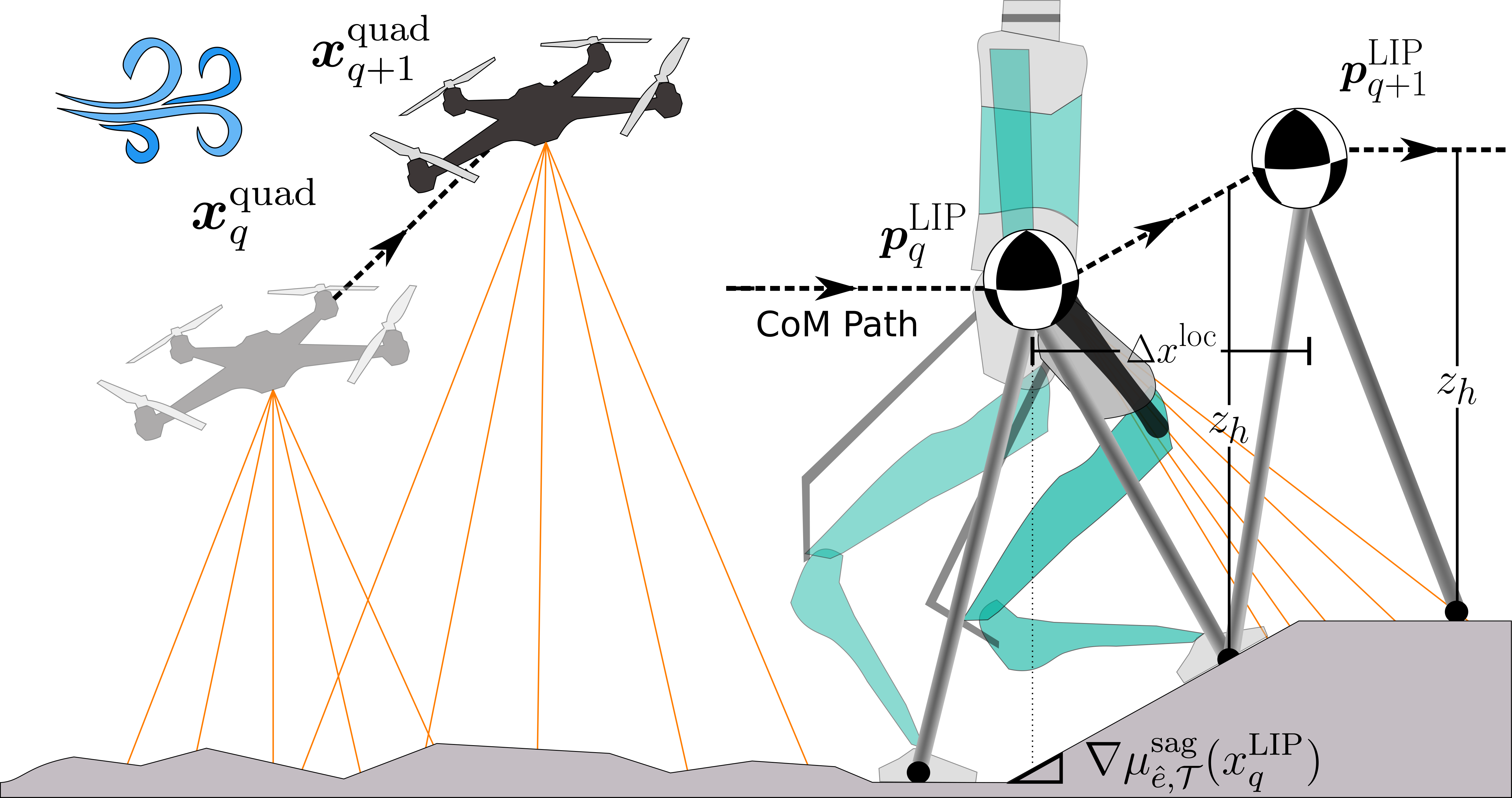}}
\caption{Illustration of the LIP model for bipedal locomotion and the quadrotor model for two consecutive discrete states. The orange lines indicate the range sensors attached to the robots and an illustration of the corresponding local field of view for measuring the terrain data.}
\label{fig:digit_lip}
\vspace{-0.2in}
\end{figure}

We leverage a locomotion-specific reduced order model (ROM) to describe our Digit bipedal robot dynamics in this study. The ROM used to design the 3-D walking motion is the linear inverted pendulum (LIP) model \cite{kajita20013d}. For the LIP model we assume that each walking step has a fixed time duration $D$~\cite{narkhede2022sequential, teng2021toward, shamsah2024socially, gibson2022terrain}.
Then, we build our model on the discrete sagittal dynamics\footnote{The lateral dynamics are only considered in the ALIP model at the low level since they are periodic with a constant desired lateral foot placement.} $(\Delta x\local \currq,v\local\currq)$, where $x\local\currq$ is the CoM position at the beginning of the $q^{\rm th}$ step, $\Delta x_q\local=x\local\nextq - x\local\currq$ is the local sagittal CoM position difference between two consecutive walking steps\footnote{The robot model used in our study represent step-by-step dynamics, i.e., $\state\currq\lip$ and $\state\nextq\lip$ are the CoM state at the foot contact switching instant of two consecutive walking steps.}, and $v\local\currq$ is the sagittal velocity at the local coordinate for the $q^{\rm th}$ walking step, similar to~\cite{shamsah2024socially}. The 3-D LIP dynamics become:
\begin{subequations}
\label{eq:lip_dynamics}
\begin{align}
x\nextq\lip &= x_q\lip +\Delta x_q\local (u^f\currq)\cos(\theta_q) \\
y\nextq\lip &= y_q\lip + \Delta x_q\local (u^f\currq) \sin(\theta_q) \\
z\nextq\lip &= z\currq\lip + \nabla\mu^{\rm sag}_{\hat{e},\set{T}}(x_q\lip) \Delta x\local\\
v\local\nextq  &= \cosh(\beta D) v\local\currq - \beta \sinh(\beta D) u^f\currq\\
\theta\lip\nextq &= \theta\currq + u^{\Delta \theta}\currq 
\end{align}
\end{subequations}

 where $u^{f}_{q}$ is the sagittal foot position relative to the CoM, $u^{\Delta \theta} \currq$ is the heading angle change, $\beta = \sqrt{g/z_h}$, where $g$ is the gravitational constant, $z_h$ is the constant CoM height at apex state\footnote{Apex state is the state when the CoM is directly on top of the foot.} (see Fig.~\ref{fig:digit_lip}), $\nabla\mu^{\rm sag}_{\hat{e},\set{T}}(x_q\lip)$ is the terrain slope at the current step (see Fig.~\ref{fig:digit_lip}), and $\theta$ is the heading. For notational simplicity, (\ref{eq:lip_dynamics}) will hereafter be referred to as:
\begin{equation}
    \state\lip \nextq = \Phi\lip(\state\lip \currq, \ctrl\lip \currq)
    \label{eq:ROM_general}
\end{equation}
where $\state_q\lip=(\position_q \lip, v_q\local, \theta_q\lip)$ and $\position_q \lip = (x_q\lip,y_q\lip, z_q\lip)$ is the 3-D location in the world coordinate at step $q$. The control input is $\ctrl \currq \lip= (u^{f}_{q},  u^{\Delta \theta} \currq)$. A detailed derivation of~\eqref{eq:lip_dynamics} is in~\cite{shamsah2024socially}.


\subsection{Quadrotor Model}
\label{subsec:quad_model}
We use a 10-D quadrotor model assuming near-hover conditions (\textit{i.e.}, small pitch and roll angles)~\cite{chung2024goal, chen2021fastrack} and consider the discrete dynamics as:
\begin{equation}
    \state\quadrotor \nextq = \Phi\quadrotor(\state\quadrotor \currq, \ctrl\quadrotor \currq)
    \label{eq:quad_general}
\end{equation}
where $\state\quadrotor \currq=(\position\quadrotor_q, \boldsymbol{v}\quadrotor_q, \boldsymbol{\theta}\quadrotor_q,\boldsymbol{\omega}_q)$, $\position_q\quadrotor=(x\quadrotor\currq,y\quadrotor\currq,z\quadrotor\currq)$ denotes the position at the $q^{\rm th}$ time step, $\boldsymbol{v}_q\quadrotor=(v\quadrotor_{x,q},v\quadrotor_{y,q},v\quadrotor_{z,q})$ are velocities, $\boldsymbol{\theta}_q\quadrotor=(\theta\quadrotor_{x},\theta\quadrotor_{y})$ are pitch and roll, and $\boldsymbol{\omega}_q=( \omega_{x,q},  \omega_{y,q})$ are pitch and roll rates. 
The control input $\ctrl_q\quadrotor = (\alpha_{x,q}, \alpha_{y,q}, \alpha_{z,q})$ is the desired pitch and roll $(\alpha_{x,q}, \alpha_{y,q})$ and the vertical thrust $\alpha_{z,q}$. The quadrotor dynamics are discretized with the same step time $D$ as the LIP model.

We refer to~\cite{chen2021fastrack} for a detailed description of the dynamics.

\subsection{Problem Statement}

\begin{prob}
Given the bipedal robot and quadrotor dynamics in Sec. \ref{sec:prob_formulation}, along with terrain elevation data $\set{T}$ and belief points $\set{B}$,  design a task allocation and assignment algorithm to search for and rescue the subjects $S_i, \; \forall i \in K$. Synthesize an optimal motion plan which leverages the distinct capabilities of the robots while ensuring safety.
\label{problem_statment}
\end{prob}

%% file: sections/03_terrainGP.tex
\section{Gaussian Process Learning of Terrain and Target Belief}
\subsection{Gaussian Processes}\label{subsection: GP Definition}
To learn the terrain and belief uncertainties present in our environment, we use Gaussian process (GP) regression:
\begin{defn}[Gaussian Process Regression]
\label{def:GP}%
Gaussian Process (GP) regression models a function $g_i:\mathbb{R}^n\to \mathbb{R}$ as a distribution with covariance $\kappa:\mathbb{R}^n\times\mathbb{R}^n\xrightarrow{}\mathbb{R}_{>0}$. Assume a dataset of $m$ samples $\mathcal{D} = \{(\bm{\xi}^j,\lambda_i^j)\}_{j\in\{1,...,m\}}$, where $\bm{\xi}^j\in\mathbb{R}^n$ is the input and $\lambda^j_i$ is an observation of $g_i(\bm{\xi}^j)$ under Gaussian noise with variance $\sigma_{\nu_i}^2$. 
Let $K\in \mathbb{R}^{m\times m}$ be a kernel matrix defined elementwise by $K_{j\ell}=\kappa(\bm{\xi}^j,\bm{\xi}^\ell)$ and for $\bm{\xi}\in\mathbb{R}^n$, let $k(\bm{\xi})=[\kappa(\bm{\xi},\bm{\xi}^1) \;  \kappa(\bm{\xi},\bm{\xi}^2) \ldots $ $\kappa(\bm{\xi},\bm{\xi}^m)]^T\in \mathbb{R}^m$.
Then, the predictive distribution of $g_i$ at a test point $\bm{\xi}$ is Gaussian with mean $\mu_{g_i,\mathcal{D}}$ and variance $\sigma_{g_i,\mathcal{D}}^2$ given by
\begin{align}
    \nonumber\mu_{g_i,\mathcal{D}}(\bm{\xi}) &= k(\bm{\xi})^T(K+\sigma_{\nu_i}^2I_m)^{-1}\Lambda\\
    \nonumber\label{Std Deviation} \sigma_{g_i,\mathcal{D}}^2(\bm{\xi})&=\kappa(\bm{\xi},\bm{\xi})-k(\bm{\xi})^T(K+\sigma_{\nu_i}^2I_m)^{-1}k(\bm{\xi}),
\end{align}
where $I_m$ is the identity and $\Lambda=\begin{bmatrix}\lambda^1_i& \lambda^2_i & \ldots & \lambda^m_i\end{bmatrix}^T$.
\end{defn}

In this work, we use the radial basis function (RBF) kernel
\begin{equation}
   \label{eg:rbf}
   \kappa(\bm{\xi}^i,\bm{\xi}^j) = \sigma_{f}^2 \exp\left(-\|\bm{\xi}^i - \bm{\xi}^j\|^2 / 2\ell^2 \right),
\end{equation}
where \( \sigma_{f}^2 \) and \( \ell \) are the signal variance and lengthscale hyperparameters, respectively. 
We will also use the partial derivative \cite{johnson2020kernel} with respect to $\xi_{i'}^i$, the dimension $i'$ of $\bm{\xi}^i$:
\begin{equation}\label{eq:partial_derivative}
\frac{\partial\kappa(\boldsymbol{\xi}^i,\boldsymbol{\xi}^j)}{\partial \xi_{i'}^i} = \frac{-\sigma_{f}^2}{\ell^2}({\xi}_{i'}^i-{\xi}_{i'}^j)\kappa(\boldsymbol{\xi}^i,\boldsymbol{\xi}^j).
\end{equation}


\subsection{Terrain GP for Locomotion Safety}\label{subsec:terrain_gp}

We use the terrain elevation data $\set{T} = \{ (\position_i, e_i)\}^{n}_{i=0}$ to train a terrain GP $\hat{e}(\position)$ for which the input is a global location $(x,y)$ and the output is the predicted terrain height at that location, $\hat{e}$. Given $\hat{e}(\position)$ and assuming no sensor noise, the mean of the terrain elevation at some test point $\position^t$ is
\begin{equation}\label{eq:terrain_mean}
    \mu_{\hat{e},\set{T}}(\position^t) = k(\position^t)^T K^{-1}E,
\end{equation}
where $E=\begin{bmatrix}e_1 & e_2 & \ldots & e_n\end{bmatrix}^T$ is the elevation data. 

As described in \cite[Chap. 9]{Williams96gaussianprocesses}, given the terrain elevation data $\set{T}$ we can construct a slope GP by taking the gradients of \eqref{eq:terrain_mean} with respect to the test point in the $x$ and $y$ dimensions:
\begin{equation}\label{eq:terrain_slope}
\nabla\mu_{\hat{e},\set{T},\{x,y\}}(\position^t)=\nabla_{\position^t,\{x,y\}}k(\position^t)^T K^{-1}E,
\end{equation}
where $\nabla_{\position^t,\{x,y\}}k(\position^t)$ is a vector of the partial derivative of $k(\position^t)$ with respect to $\position^t$, constructed by applying~\eqref{eq:partial_derivative} elementwise with respect to either the $x$ or $y$ dimension.

We then transform the gradients to the sagittal and lateral directions relative to the current robot heading:

\begin{equation}\label{eq:local_slope}
    \left[\nabla\mu^{\rm sag}_{\hat{e},\set{T}}, \nabla\mu^{\rm lat}_{\hat{e},\set{T}} \right]^\top = R(\theta\lip) \cdot \left[\nabla\mu_{\hat{e},\set{T},x} , \nabla\mu_{\hat{e},\set{T},y}\right]^\top
\end{equation}
where $R(\theta\lip)$ is a rotation matrix based on the LIP modeling heading angle. In Sec.~\ref{sec:mpc},~\eqref{eq:local_slope} will be minimized in our proposed MPC to encourage the walking robot to choose a path with low terrain slopes.

\subsection{Belief GP for Search and Rescue Task}\label{subsec:belief_gp}

We use the initial belief data $\set{B} = \{ (\position^s_i, c^s_i)\}^{m}_{i=1}$ to generate training points $\hat{\set{B}} = \{\hat{\set{B}_i}\}_{i=1}^m$, where $\hat{\set{B}}_i$ is a set of $l$ points sampled from the normal distribution $\{\mathcal{N}(\position^s_i,  {c^s_i}^2)\}$. 
We then train a belief GP $\hat{b}(\position)$ for which the input is a global potential location $(x,y)$ of the subject and the output is the predicted belief at that location, $\hat{b}$. 
We solve for the mean of the belief value at point $\position^t$ as $ \mu_{\hat{b},\set{B}}(\position^t) = k(\position^t)^T K^{-1}\hat{\set{B}}$.

\section{Multi-Robot Terrain-Aware MPC}\label{sec:mpc}

In this section, we design a terrain-aware motion planner for the heterogeneous multi-robot team. Compared to our previous works on integrating terrain GPs with locomotion planners, the objective of this planner is not to avoid terrains with high terrain uncertainty~\cite{jiang2023abstraction} or elevations~\cite{muenprasitivej2024bipedal}. Instead, we aim to generate safe paths for the bipedal robot to traverse rough terrain while minimizing the lateral slopes of the planned paths using gradient-based trajectory optimization.

To this end, we integrate lateral slopes derived from the terrain GP in~\eqref{eq:local_slope} into the cost function in our proposed MPC framework. The terrain GP learns a smooth, differentiable, and continuous approximation of the rough terrain. 
This makes it an ideal candidate for gradient-based motion planning methods as proposed here:
\begin{subequations}
\label{general_mpc}
\begin{align*}
\min_{X, U} \quad \sum_{q=0}^{N-1} & J(\boldsymbol{x}\lip_q, \boldsymbol{x}\quadrotor_q, \boldsymbol{u}\lip_q, \boldsymbol{u}\quadrotor_q) +  \nabla\mu^{\rm lat}_{\hat{e},\set{T}}(\position\lip_q)^2 \\
\textrm{s.t.} \quad & \state\lip \nextq = \Phi\lip(\state\lip \currq, \ctrl\lip \currq) \\
    & \state\quadrotor \nextq = \Phi\quadrotor(\state\quadrotor \currq, \ctrl\quadrotor \currq)\\
    & \boldsymbol{x}\lip_0 = \boldsymbol{x}\lip_{\rm init}, \;  (\boldsymbol{x}\lip_q, \boldsymbol{u}\lip_q) \in \mathcal{XU}\lip_q  \\
    & \boldsymbol{x}\quadrotor_0 = \boldsymbol{x}\quadrotor_{\rm init}, \;  (\boldsymbol{x}\quadrotor_q, \boldsymbol{u}\quadrotor_q) \in \mathcal{XU}\quadrotor_q \\
    & d_{l} - \epsilon \leq d(\position\lipone\currq, \position\liptwo\currq) \leq d_{u} + \epsilon 
\end{align*}
\end{subequations}
where $J(\state, \ctrl)$ is a cost that penalizes the distance between the current CoM state of each robot and its corresponding target location and $\nabla\mu^{\rm lat}_{\hat{e},\set{T}}(\position\lip_q)^2$ is used to minimize the lateral slope of the path that the LIP model takes to reach the target location. $X$ is the state vector of all the robots $(\state\lip,\state\quadrotor)$\footnote{Note that the superscripts $(\cdot)^{\lip}$ and $(\cdot)^{\quadrotor}$ refer to both humanoid LIP models and both quadrotors unless otherwise noted.}, and $U$ is the control vector $(\ctrl\lip, \ctrl \quadrotor)$. $\mathcal{XU}\lip$ and $\mathcal{XU}\quadrotor$ are linear constraints on the states and control inputs for the robots. For the legged robots not to collide with each other, we enforce a distance constraint between the robots, where $d(\position\lipone\currq, \position\liptwo\currq)$ is the Euclidean distance between the two robots, and $(d_{l},d_{u})$ are upper and lower bounds, respectively, on the allowed distance\footnote{The upper bound constraint is only active during bipedal robot rescue tasks (see Sec.~\ref{subsec:rescue_task}).}.
This MPC is used as the underlying motion planner for all the tasks detailed in Sec.~\ref{sec:task_assignment}. The MPC formulation is general and can be generalized to simultaneously solve the paths for a team of multiple humanoids and quadrotors. 

%% file: sections/04_taskplanner.tex
\section{Task Planner}
\subsection{Task Specification}
In this work, we use the semantics of syntactically cosafe Linear Temporal Logic (scLTL) to define tasks.
\begin{defn}[Syntactically co-safe LTL {\cite[Def. 2.3]{belta_formal_2017}}]
A \emph{syntactically co-safe linear temporal logic (scLTL)} formula $\phi$ over a set of observations $O$ is recursively defined as 
\begin{equation*}
    \phi = \true \ |\ o \ |\ \lnot{o} \ |\ \phi_1 \land \phi_2 \ |\ \phi_1 \lor \phi_2 \ |\ \lnext \phi \ |\ \phi_1\until\phi_2 \ |\ \eventually \phi \ |\ \phi_1\rightarrow\phi_2
\end{equation*}
where $o\in O$ is an observation and $\phi$, $\phi_1$, and $\phi_2$ are scLTL formulas.
We define the \textit{next} operator $\lnext$ as meaning that $\phi$ will be satisfied in the next state transition, the \textit{until} operator $\until$ as meaning that the system satisfies $\phi_1$ until it satisfies $\phi_2$, the \textit{eventually} operator $\eventually$ as $\true\until\phi$, and the \textit{implication} operator $\rightarrow$ as $\lnot\phi_1\lor\phi_2$.
\end{defn}
Satisfaction of scLTL is checked with finite state automata.

\begin{defn}[Finite State Automaton {\cite{belta_formal_2017}}]
A \emph{finite state automaton (FSA)} is a tuple $\mathcal{A} = (S,s_0,O,\delta,F)$, where
\begin{itemize}
    \item $S$ is a finite set of states,
    \item $s_0 \in S$ is the initial state,
    \item $O$ is the input alphabet,
    \item $\delta:S\times O\xrightarrow{} S$ is a transition function, and
    \item $F \subseteq S$ is the set of accepting (final) states.
\end{itemize}
\end{defn}

We define the observation space $O$ as a combination of \textit{robot-} and \textit{environment-centric} observations.
\begin{defn}[Robot-centric Observations]
    Robot-centric observations $O_R$ solely depend on actions taken by robotic agents, where we have:
    \begin{itemize}
        \item $O_{R1}$: Found subject $n$, 
        \item $O_{R2}$: Rescued subject $n$ with a quadcopter,
        \item $O_{R3}$: Rescued subject $n$ with a bipedal robot,
    \end{itemize}
\end{defn}
\begin{defn}[Environment-centric Observations]
    Environment-centric observations $O_E$ depend on the state of an environmental condition, where we have:
    \begin{itemize}
        \item $O_{E1}$: Wind - quadcopter unable to rescue a subject
        \item $O_{E2}$: Untraversable terrain - bipedal robot unable to rescue a subject 
    \end{itemize}
\end{defn}

We define an individual scLTL specification for each subject requiring a rescue, 
Thus, for a scenario with $N$ subjects we concurrently use $N$ FSAs.

\subsection{Multi-robot Task Allocation}\label{subsec:task_allocation}

We now detail a high-level planning and task allocation policy. We first define robot tasks.
\begin{defn}[Robot Task]
    A robot task is a tuple $\Gamma_i=(\underline{n},\overline{n},\texttt{NAME},\texttt{ROBOT})$, where $\underline{n},\overline{n}$ are lower and upper bounds, respectively, on the number of agents required for the task, $\texttt{NAME}$ denotes the task name,
    and $\texttt{ROBOT}$ denotes the types of robots eligible for the task. 
    In this work, we have the set $\Gamma$ of robot tasks
    \begin{itemize}
        \item (Search) $\Gamma_1$: (1,$\infty$,\texttt{SEARCH},\{\texttt{QUADROTOR},\texttt{BIPEDAL}\}),
        \item (Quadrotor rescued) $\Gamma_2$: (1,1,\texttt{RESCUE},\texttt{QUADROTOR}),
        \item (Bipedal robot rescued) $\Gamma_3$: (2,2,\texttt{RESCUE},\texttt{BIPEDAL}),
        \item (Mapping terrain) $\Gamma_4$: (1,1,\texttt{MAPPING},\texttt{QUADROTOR}).
    \end{itemize}
\end{defn}
Then, we create a mapping $\mathcal{M}:O_R \rightarrow \Gamma$ from robot-centric observations to a set of robot tasks. In this work, we use the following mappings
\begin{equation*}
    O_{R1}\rightarrow\Gamma_1, 
    O_{R2}\rightarrow\Gamma_2, 
    O_{R3}\rightarrow\{\Gamma_3,\Gamma_4\}
\end{equation*}

We now detail the entire planning framework. We first define a scLTL specification for each mission subject and generate a corresponding FSA.
We assume that at each state in any FSA, there is only one robot-centric observation available. 
For each FSA representing a subject, we query the robot-centric observation available at the current state and map it to its corresponding robot tasks. These are then sent to a task assignment algorithm detailed in Section \ref{sec:task_assignment}. The agents execute their assignments until a robot-centric observation is returned from any of the agents. 
The corresponding FSA is then updated to a new state based on the robot- and environment-centric observations.
Finally, tasks are reassigned for all agents.

%% file: sections/05_MPC.tex
\section{Search and Rescue Task Assignment}\label{sec:task_assignment}

The proposed search and rescue problem consists of three subtasks (search, rescue, and mapping), as discussed in the previous section, that guide the robot team in successfully transporting the subjects to a safe location $R$, known \textit{a priori}. We now detail the metrics used to assign these subtasks.

\subsection{Search Task}\label{subsec:search}
For the search task, we choose a consensus-based auction algorithm~\cite{choi2009consensus} as it balances computational efficiency and solution quality (\textit{i.e.}, optimality) effectively, particularly when considering heterogeneous robot capabilities~\cite{badreldin2013comparative}. Auction-based methods, although computationally efficient, can lead to suboptimal task allocations, especially in scenarios with complex dependencies and path conflicts\cite{quinton2023market}. Consensus-based methods leverage the initial efficiency of auction-based allocation to quickly assign tasks based on overall scores, and then employ a consensus phase to iteratively resolve task conflicts~\cite{choi2009consensus}. In our case, task conflict arises when the planned paths for robots to complete their assigned task intersect, resulting in redundant environmental exploration. Here, instead of using a task bundle similar to the traditional consensus-based methods~\cite{choi2009consensus}, made up of all the intermediate targets explored along the path, we use line segment intersection to indicate task conflict.

\subsubsection{Auction Phase} 
\begin{figure}
    \centering
    \includegraphics[width=0.8\linewidth]{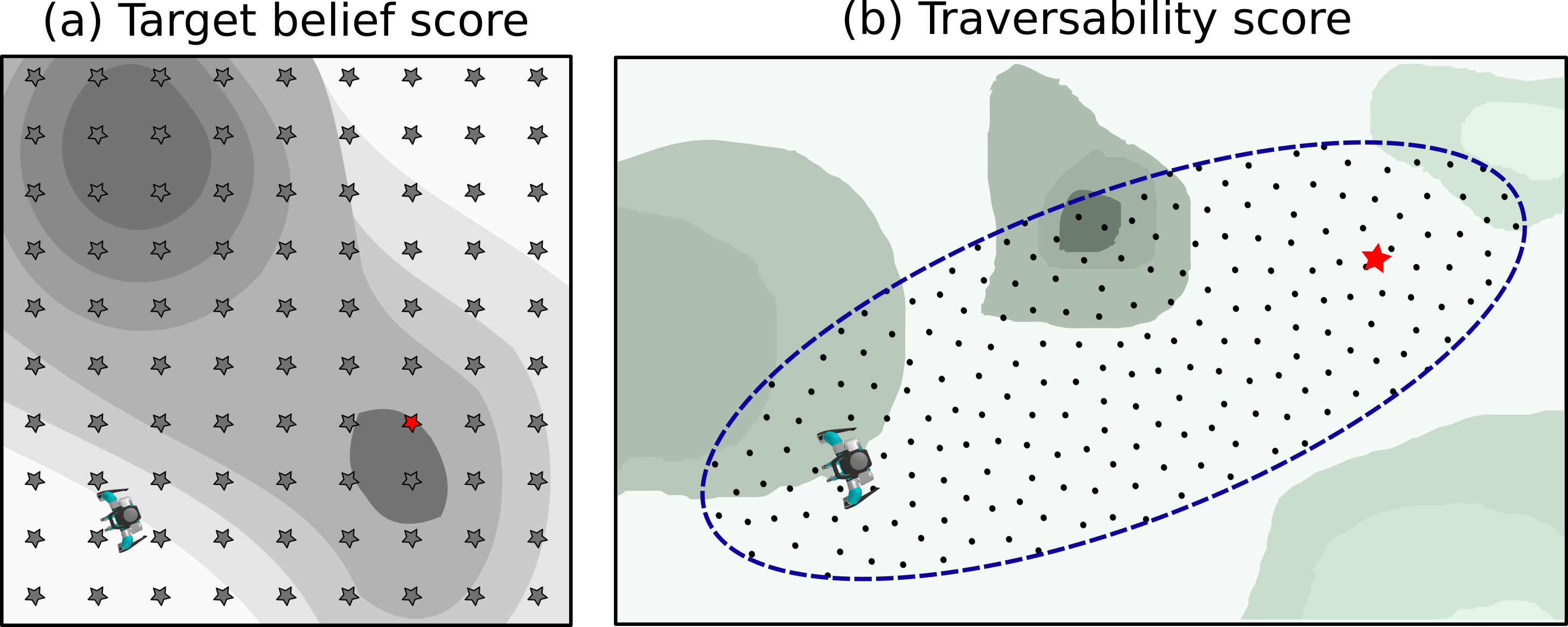}
    \caption{(a) shows the candidate target points $\boldsymbol{t}$ as $\star$, based on the belief GP of the environment shown as the gray-white gradients. (b) shows the sample points for the calculation of the traversability score $\position^{\rm ellipse}_i, \; i \in M $, as the black dots inside the ellipse. The ellipse focal points are the current position of the robot and the candidate target point shown as a red $\star$. The terrain GP of the environment is shown by the green-white gradient. }
    \label{fig:score}
    \vspace{-0.2in}
\end{figure}

In this phase each robot "bids" on the available tasks (targets in the environment to search) based on its own capabilities~\cite{quinton2023market}. We consider a set of $T$ uniformly distributed candidate target points $\{\boldsymbol{t}_i\in\mathbb{R}^2\}_{i=1}^T$ as shown by the $\star$ in Fig.~\ref{fig:score}(a). In the proposed scenario, the robots score the targets based on the belief value, the traversability, and the time it takes to reach the target.

\paragraph{Belief score $\set{S}_b$}

 represents the probability that the subject is located at a specific point. This score is derived from the belief GP model $\hat{b}(\position)$, which predicts the likelihood of the subject's presence based on observed data. Mathematically, the belief score for a target $\boldsymbol{t}_i$ 
 is given by the maximum upper confidence bound:
\begin{equation}
\set{S}_b(\boldsymbol{t}_i) = \mu_{\hat{b},\set{B}}(\boldsymbol{t}_i) +\alpha_b \sigma_{\hat{b},\set{B}}(\boldsymbol{t}_i)), \; \forall i \in [1,T]
\end{equation}

where $\sigma_{\hat{b},\set{B}}(\boldsymbol{t}_i)$ is the variance of the belief and $\alpha_b $ is a parameter that balances the mean and variance.

\paragraph{Traversability score ($\set{S}_t$)}
 evaluates the ease of traversing the terrain between the robot's current position and the target point. The traversability is evaluated based on the upper confidence bound of the terrain slope estimation at the sampled points. 
 The work in~\cite{quann2017energy} uses GP integrals to evaluate the traversability between two points. While that method provides a continuous evaluation, it is computationally expensive. As the proposed motion planner in Sec.~\ref{sec:mpc} will solve for an optimal path with high traversability (\textit{i.e.}, minimum lateral slopes), a continuous estimation of the slopes is not required, and sampling the maximum upper confidence bound of a set of randomly selected discrete points is sufficient to estimate the traversability score.
 
To compute the traversability score, we first need an initial estimation of the path. We can use a heuristic such as the straight-line path~\cite{quann2017energy} or any predefined path planning algorithm. In this work, we sample points in an ellipse shape $\position^{\rm ellipse}_i, \; i \in M $, where $M$ is the number of sampled points in the ellipse, with its focal points being the current position of the robot and a candidate target point $\boldsymbol{t}_i$ (see Fig.~\ref{fig:score}(b)). This shape models the potential traversable area more realistically, as the legged robot will not always move along a straight line to the target point due to the potential rough terrain it encounters.
We solve for the mean of the slope prediction for each point inside the ellipse by evaluating the GP $\nabla\mu_{\hat{e},\set{T},\{x,y\}}(\position^{\rm ellipse}_i), \; i \in M$. 

The traversability score is then calculated as:
\begin{equation}
\set{S}_t(\boldsymbol{t}_i) = -\mu_{\rm ellipse}(\boldsymbol{t}_i) - \alpha_t \sigma_{\rm ellipse}(\boldsymbol{t}_i), \; \forall i \in [1,T]
\end{equation}
where $\mu_{\rm ellipse}(\boldsymbol{t}_i)$ and $\sigma_{\rm ellipse}(\boldsymbol{t}_i)$ represent the average and the standard deviation of the $M$ slopes inside the ellipse for each candidate target point $\boldsymbol{t}_i$. The parameter $\alpha_t$ balances the mean and variance.

\paragraph{Time score $\set{S}_d$} accounts for the distance between the robot's current position and the target point, scaled by the robot's maximum velocity. This score ensures that a target location closer to the robot is prioritized. The time score is calculated as:
\begin{equation}
\set{S}_d(\boldsymbol{t}_i) = -\|\boldsymbol{t}_i  - \position_{0}\| / v_{\text{max}}
\end{equation}
where $\position_{0}$ is the robot's current position, and $v_{\text{max}}$ is the robot's maximum velocity.

The total score $S_{\text{total}}$ for each point is a weighted sum of the individual scores:
\begin{equation}
\set{S}_{\text{total}}(\boldsymbol{t}_i) = w_b \set{S}_b(\boldsymbol{t}_i)  + w_t \set{S}_t(\boldsymbol{t}_i)
+ w_d \set{S}_d(\boldsymbol{t}_i)\end{equation}
where $w_b$, $w_t$, and $w_d$ are the nonnegative weights assigned to the belief, traversability, and time scores, respectively. These weights can be tailored based on the task priorities.

\subsubsection{Consensus Phase}\label{subsubsec:consensus} Solely relying on the auction phase for task allocations will potentially lead to task conflicts between the robots, where individual robots have the same targets or their paths to their target locations explore the same intermediate points. In the consensus phase, we propose a method to resolve such conflicts, by iterating through the $T$ target locations of the two conflicting robots and selecting the targets that lead to the highest sum of scores for both robots without conflicts. As an example, consider a scenario in which $\position\lip$ has a target $\boldsymbol{t}\lip_{1}$ and $\position\quadrotor$ has a target $\boldsymbol{t}\quadrotor_{1}$, and they are in conflict. The conflict resolution solves for the optimal targets $(\boldsymbol{t}\lip_{i},\boldsymbol{t}\quadrotor_{j})$ such that the sum $\left( \set{S}_{\text{total}}(\boldsymbol{t}\lip_i) + \set{S}_{\text{total}}(\boldsymbol{t}\quadrotor_j)  \right)$ is maximized and no conflict exists for the newly assigned targets.


\subsection{Rescue Task} \label{subsec:rescue_task}
The rescue task consists of transferring the subjects $S_1$ and $S_2$ to the rescue location $R$. In the targeted scenario we set that a quadrotor can rescue $S_1$, while $S_2$ requires two legged robots for the rescue mission.
\subsubsection{Rescue of $S_1$ with A Quadrotor} 
Once $S_1$ is found, the target location for the nearest quadrotor (e.g., $q_2$) is set to reach $S_1$. Once a quadrotor reaches  $S_1$, the target location for  $q_2$ is set to $R$. While $q_2$ is rescuing $S_2$, $q_1$ is continuously searching for $S_2$ with the bipedal robot team.

\begin{figure*}[t]
\centerline{\includegraphics[width=.88\textwidth]{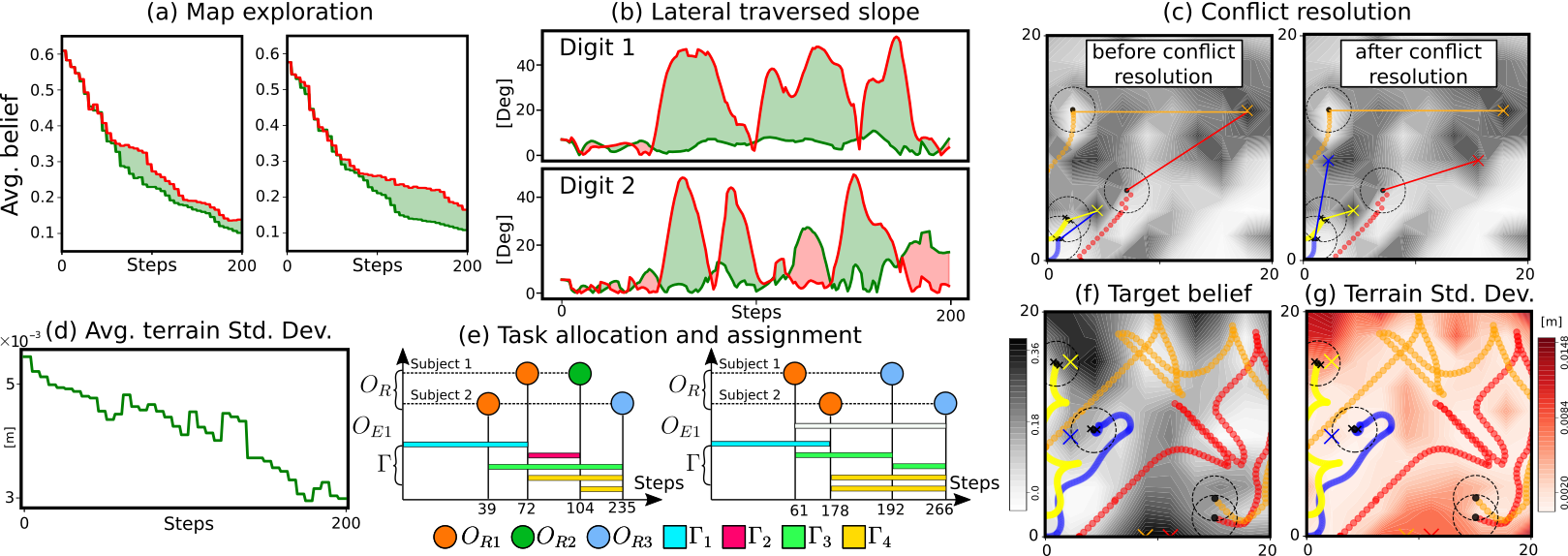}}
\caption{Planning results: (a) shows the average belief value for two different trials with randomly selected initial belief conditions $\set{B}$. The red line is trial without conflict resolution and the green line is with conflict resolution. (b) shows the lateral slopes traversed by both Digits. The red line is a trial without traversiblity score and slope minimization in the MPC and the green line is with traversiblity score and slope minimization. (c) shows conflict resolution when the auction phase outputs the same targets for the robot team, where the straight lines connect robots to their target points, (d) shows the average standard deviation of the terrain GP, (e) shows the task allocation and assignment results of two different runs with and without wind $O_{E1}$, (f) shows the belief value, and (g) shows that the standard deviation of the terrain estimation decreases as the robots explore the environment.}
\label{fig:comb_results}
\vspace{-0.2in}
\end{figure*}

\subsubsection{Rescue of $S_2$ with Two Bipedal Robots} 
Once $S_2$ is found both legged robot targets are set to $S_2$. The legged robots are tasked to maintain a specific distance between them and carry $S_2$ to $R$. In this phase of the task, the available quadrotors (e.g. quadrotor that is not rescuing $S_1$) continuously map the terrain around the two bipeds.


\subsection{Mapping Task}
All robots update the terrain GP based on current measurements of the surrounding environment during the searching task. However, during the rescue phase, the quadrotors play a critical role in mapping the potential path that the legged robots will take. The quadrotors are tasked to survey the terrain by orbiting around the bipedal robots:
\begin{equation*}
    \boldsymbol{t}\quadrotor = \left( x\nextq\lip + r \cdot \cos\left(\frac{2 \cdot s}{\pi}\right), \, y\nextq\lip + r \cdot \sin\left(\frac{2 \cdot s}{\pi}\right) \right)
\end{equation*}
where $r$ is the radius of the circle around the biped CoM state and $s$ is the step increment. 


%% file: sections/07_results.tex
\begin{table}[t]
    \centering
    \caption{Parameter Values of Our Results}
    \begin{tabular}{|c|c|c|c|}
        \hline
        \textbf{Parameter} & \textbf{Value} & \textbf{Parameter} & \textbf{Value} \\ 
        \hline
        Environment & $20 \times 20$ m$^2$ & $R$ & $(19, 19)$ m \\ 
        $N$ & $10$ & $D$ & $0.4$ s \\ 
        $d_l$ & $0.9$ m & $d_u$ & $2$ m \\ 
        $\epsilon$ & $0.2$ m & $\alpha_b$ & $3$ \\ 
        $T$ & $100$ & $M$ & $15$ \\ 
        $\alpha_t$ & $1$ & $w_b$ & $3$ \\ 
        $w_d\quadrotor$  & $0.5$ & $w_d\lip$ & $1$ \\ 
        $w_t\lip$ & $1$ & $v_{\rm max}\quadrotor$ & $1.5$ m/s \\ 
        $v_{\rm max}\lip$ & $0.4$ m/s & $r$ & $3$  \\ 
        \hline
    \end{tabular}
    \label{tab:param}
    \vspace{-0.2in}
\end{table}

\section{Implementation and Results}
In this section, we will present the results of the proposed framework, in particular, the searching task and the terrain-aware MPC to minimize lateral slope for humanoid robot trajectories\footnote{For rescue and mapping task please see the video~\href{https://www.youtube.com/watch?v=yWMuhZYh1HI}{\nolinkurl{https://www.youtube.com/watch?v=yWMuhZYh1HI}.}}. Details of the implementation parameters are shown in Table~\ref{tab:param}.

\subsection{Task Specification}
Here we show the scLTL specifications used for the search and rescue task.

\textit{Subject 1: } This specification indicates that subject 1 can only be rescued by the quadrotor when there is no wind at the location ($\lnot O_{E1}$) and can be rescued by the bipedal robot when wind exists and the terrain is traversable ($O_{E1}\land \lnot O_{E2}$):
\begin{align*}
    \eventually O_{R1}\land &[([O_{R1}\land\lnot O_{E1}]\rightarrow\lnext O_{R2}) \\
    &\lor([O_{R1}\land O_{E1}\land \lnot O_{E2}]\rightarrow\lnext O_{R3})]
\end{align*}

\textit{Subject 2: } This specification indicates that subject 2 can only be rescued by the bipedal robots:
\begin{equation*}
    \eventually O_{R1} \land [(O_{R1}\land \lnot O_{E2})\rightarrow\lnext O_{R3}]
\end{equation*}

Fig. \ref{fig:comb_results}(e) depicts the progress through the scLTL specifications along with the corresponding task sequences for two sample runs of the mission.

\subsection{Map Exploration and Conflict Resolution}
We demonstrate the map exploration performance by inspecting the average target belief (Fig.~\ref{fig:comb_results}(a)) of the environment with and without conflict resolution (Fig.~\ref{fig:comb_results}(c)).
As shown in Fig.~\ref{fig:comb_results}(a), in two different trials with different random initial target belief information $\set{B}$, the average belief value of the environment decreases as the robots explore the environment. We specifically show that with conflict resolution (green line) the average belief reduces at a faster rate than without conflict resolution (red line).
Fig.~\ref{fig:comb_results}(f), shows the belief value in the environment, and specifically, the reduction of belief uncertainty as the robots navigate through the environment. 

\subsection{Traversibility and Terrain-Aware MPC}
In Fig.~\ref{fig:comb_results}(b), we show the result of integrating the terrain GP into the MPC to solve for paths with minimal lateral slopes, as well as adding the traversability score into the task allocation. Our proposed framework produces paths with low lateral slopes (green line) compared to the lateral slopes in the paths produced when not including terrain slope into the MPC and the traversability score in the total score (red line). Traversing paths with lower lateral slopes increases the locomotion safety of the humanoids as lateral stability is critical to maintaining balance~\cite{gu2024robust,gibson2022terrain}. In Fig.~\ref{fig:comb_results}(d) we show the standard deviation of the terrain GP, indicating the uncertainty reduction as the robots explore the environment. The average standard deviation decreased by $\approx 45 \%$ in $200$ steps. 
Fig.~\ref{fig:comb_results}(g) shows that the standard deviation decreases along the explored paths.

%% file: sections/08_conclusion.tex
\section{Conclusion}

This study presents a terrain-aware MPC for heterogeneous bipedal and aerial multi-robot teams conducting search and rescue missions in uncertain environments. 
Future work will focus on (i) expanding the task set to the full semantics of scLTL beyond the search and rescue domain, and (ii) potential hardware implementations like those shown in our previous work of \cite{cao2022leveraging}.